\documentclass[conference]{IEEEtran}
\IEEEoverridecommandlockouts
\usepackage{cite}
\usepackage{amsmath,amssymb,amsfonts}
\usepackage{algorithmic}
\usepackage{graphicx}
\usepackage{textcomp}
\usepackage{tabularx}
\usepackage{booktabs}
\usepackage{xcolor}
\usepackage{array}
\def\BibTeX{{\rm B\kern-.05em{\sc i\kern-.025em b}\kern-.08em
    T\kern-.1667em\lower.7ex\hbox{E}\kern-.125emX}}
\begin{document}

\title{Robust Uncertainty Quantification for Factual Generation of Large Language Models\\
\thanks{*Zhongliang Yang is the corresponding author of this paper. And our code is available at https://github.com/EdwardChang5467/robust\_uncertainty.}
}

\author{\IEEEauthorblockN{Yuhao Zhang, Zhongliang Yang*, Linna Zhou}
\IEEEauthorblockA{\textit{School of Cyberspace Security, 
Beijing University of Posts and Telecommunications, Beijing, China} \\
\{yuhaozhang, yangzl, zhoulinna\}@bupt.edu.cn}
}

\maketitle

\begin{abstract}
The rapid advancement of large language model (LLM) technology has facilitated its integration into various domains of professional and daily life. However, the persistent challenge of LLM hallucination has emerged as a critical limitation, significantly compromising the reliability and trustworthiness of AI-generated content. This challenge has garnered significant attention within the scientific community, prompting extensive research efforts in hallucination detection and mitigation strategies. Current methodological frameworks reveal a critical limitation: traditional uncertainty quantification approaches demonstrate effectiveness primarily within conventional question-answering paradigms, yet exhibit notable deficiencies when confronted with non-canonical or adversarial questioning strategies. This performance gap raises substantial concerns regarding the dependability of LLM responses in real-world applications requiring robust critical thinking capabilities.
This study aims to fill this gap by proposing an uncertainty quantification scenario in the task of generating with multiple facts. We have meticulously constructed a set of trap questions contained with fake names. Based on this scenario, we innovatively propose a novel and robust uncertainty quantification method(RU). A series of experiments have been conducted to verify its effectiveness. The results show that the constructed set of trap questions performs excellently. Moreover, when compared with the baseline methods on four different models, our proposed uncertainty quantification method has demonstrated great performance, with an average increase of 0.1-0.2 in ROCAUC values compared to the best performing baseline method, providing new sights and methods for addressing the hallucination issue of LLMs.
\end{abstract}


\section{Introduction}\label{I}
The extensive application of large language models (LLMs) in the field of natural language generation (NLG) has led to a growing reliance on these models in everyday life. People increasingly turn to LLMs to assist with reading and understanding documents\cite{b1}, support decision-making\cite{b2}, and complete various tasks by utilizing the models' responses and generated content. This increasing dependence has, in turn, heightened the importance of the credibility and reliability of the models' outputs. However, LLMs are inevitably prone to the issue of ``hallucination"\cite{b3}. This phenomenon, where models may produce content that is obscure or fabricated, poses a significant challenge to the credibility and reliability of the outputs. 

The hallucinations of LLMs can be categorized into factual hallucinations and faithfulness hallucinations\cite{b4}. Faithfulness hallucinations mainly evaluate whether the output is faithful to the input, while factual hallucinations primarily assess whether the generated content is consistent with reality. Faithfulness hallucinations can be identified simply by assessing the relevance between the output and the input. Factual hallucinations, characterized by their fine-grained nature and scattered distribution, are less likely to be intuitively detected\cite{b32}. Models 

\begin{figure}[htbp]
\centerline{\includegraphics[width=0.5\textwidth]{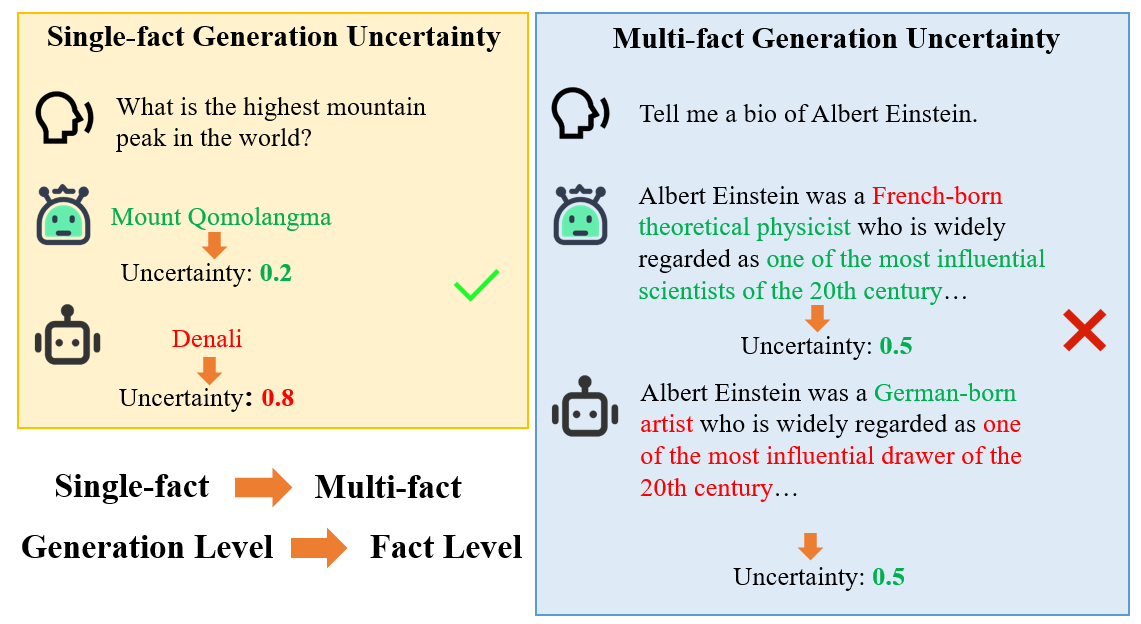}}
\caption{The difference in uncertainty quantification between single-fact generation and multi-fact generation.}
\label{fig1}
\end{figure}

\noindent may generate content that appears coherent and persuasive on the surface. For instance, in the task of generating biographies of real individuals, they may produce outputs containing wrong or fake facts. Alternatively, when inadvertently prompted by users to generate biographies of fictional individuals, the models may proceed with the task as if it were normal. Given the intractability of eliminating hallucinations in LLMs, we can address this issue by measuring the uncertainty of model-generated outputs externally. By highlighting answers with high uncertainty, we can alert users to potential inaccuracies.

Currently, several methods for quantifying the uncertainty of LLMs' generations have been proposed. However, these methods typically consider the uncertainty at the level of the entire generated text. They rely on the content of the generated text and the logits information of the generated tokens for calculation, and are primarily designed to verify single facts\cite{b33}. Nevertheless, when the generated content involves multiple facts, these existing methods still have limitations in accurately measuring the uncertainty in such cases. There can be situations where factual errors exist, but the measured uncertainty remains low, as illustrated in Fig.~\ref{fig1}.

In natural language processing (NLP), multi-fact generation involves producing text that contains multiple independent facts. Evaluating the uncertainty of multi-fact generation is more complex than single-fact generation. It requires models to break down the generated text into individual facts, quantify the uncertainty of each fact, and verify them simultaneously within complex contexts. Additionally, hallucinations in the generation process are harder to detect intuitively, further complicating the evaluation. Prior research has primarily focused on multi-fact generation based on real-world data and then conducting fact-checking based on the generated outputs \cite{b5,b6,b7,b8,b9}. However, users of LLMs may intentionally or inadvertently pose incorrect trap questions to the models in practice, such as prompting the generation of biographies for fictional individuals. In such cases, the model may either refuse to generate or, with some probability, fabricate a biography for the fictional individual or transfer facts from real biographies to the fictional one, it is not convenient to decompose the generated content into corresponding individual facts. We argue that previous methods fail to fully meet the demands of real-world users, as they lack robustness and are unable to quantify the uncertainty of generations when confronted with erroneous queries from users. This, in turn, poses a more formidable challenge for uncertainty quantification in multi-fact generation and fact verification.

Our research aims to quantify the uncertainty of multi-fact generation in LLMs based on trap questions. To this end, we have constructed a multi-fact generation scenario using trap questions. To realize this scenario, we introduce a pipeline for constructing a dataset of fictional biography generation tasks. Specifically, we created a dataset comprising 77 trap questions and 385 generations sampled from four LLMs. Moreover, we develop a robust uncertainty estimation method $RU$. Through comparisons with four models and several baseline methods, we demonstrate the superior performance of our proposed approach. To our knowledge, our work represents the first study on uncertainty estimation based on trap questions.  Our approach effectively addresses the uncertainty quantification of trap questions through fine-grained classification and quantification. We hope that our research will contribute to the study of uncertainty estimation in multi-fact generation tasks.

Our contributions can be summarized as follows:

\begin{itemize}
\item We introduce a pipeline for the construction of trap questions based on LLMs and have built a dataset $MulFactTrap$ comprising 77 trap questions based on this pipeline and included 385 generations of four LLMs based on $MulFactTrap$.
\item We propose a novel and robust white-box uncertainty estimation method for multi-fact generations of LLMs.
\item We conducted experimental validation of the proposed trap question construction method and the uncertainty measurement method. Our results demonstrate the excellent performance of these methods across four LLMs.
\end{itemize}

\section{Related work}\label{II}

The uncertainty quantification of LLM's outputs can be categorized into five types \cite{b10}: Logit-based methods \cite{b11,b12}, internal state-based methods \cite{b13,b14,b15}, verbal expression-based methods \cite{b16,b17}, consistency estimation-based methods \cite{b18,b19}, and Surrogate models-based methods \cite{b20,b21}. Logit-based methods are simple to implement and provide intuitive outputs. However, when dealing with complex nonlinear problems, they may fail to accurately capture the intricate relationships within the data, leading to underfitting. Internal state-based methods can delve into the internal space of the model but are computationally complex and highly dependent on the model architecture. Verbal expression-based methods are easy to understand but produce less accurate quantification results. Consistency estimation-based methods are applicable to both black-box and white-box models but come with high computational costs. Surrogate models-based methods offer high computational efficiency but are limited in precision.

The aforementioned studies have primarily focused on single-fact generation. In contrast, measuring the uncertainty of multi-fact generation poses greater challenges compared to single-fact generation, as it requires a finer granularity and higher accuracy. Over the past two years, researchers have proposed several methods for measuring the uncertainty of multi-fact generation. Fadeeva et al. \cite{b9} proposed the Claim-Conditioned Probability (CCP) method, which adjusts and aggregates the probabilities of tokens generated by the model to obtain uncertainty at the claim level. Jiang et al. \cite{b22} introduced the CORE component, which filters claims based on their uniqueness and informativeness generated by the model, and only conducts fact verification on the filtered facts. This approach enhances the robustness of uncertainty methods to some extent. Vazhentsev et al.\cite{b23} proposed the TAD method, which adjusts the uncertainty of the current generated token by training a regression model with the target variable being the gap between conditional and unconditional generation confidence. This method is particularly effective in long text generation but requires the construction of additional training data, resulting in a higher computational cost. However, to the best of our knowledge, previous methods have not addressed the uncertainty quantification for trap questions, nor have they studied the credibility of answers to trap questions in multi-fact generation tasks. This raises concerns about the robustness of existing uncertainty quantification methods when applied to trap questions.

\section{Robust uncertainty quantification and scenario construction}\label{III}
In this section, we will introduce the proposed uncertainty measurement scenario in Section \ref{IIIA}, and at the same time, we will introduce the method of constructing the dataset required for uncertainty quantification in this scenario in Section \ref{IIIB}.

\subsection{Robust Uncertainty Quantification Scenario Description}\label{IIIA}
The typical datasets used for uncertainty quantification\cite{b8,b9,b24} only include questions that are factually correct, lacking the inclusion of tricky or misleading questions, which is insufficiently robust for measuring the uncertainty of LLMs in practical applications and thereby assessing the credibility of their outputs. In real-life scenarios, people may pose wrong or fake questions to LLMs either intentionally (for robustness evaluation or malicious attacks) or unintentionally (due to carelessness). If the model accepts and responds to these questions based on user intent, the answers are likely to contain factual errors, and the uncertainty quantification should reflect a higher degree of uncertainty. Conversely, if the model rejects the questions or identifies the errors within them, the uncertainty quantification should yield a lower value. In this scenario, the letters and symbols used, along with their corresponding meanings, are shown in Table~\ref{tab1}.

\begin{table}[htbp]
\caption{Symbols and meanings used in the scene}
\begin{center}
\renewcommand{\arraystretch}{1.5}
\begin{tabular}{c|c}
\toprule
% \hline
\textbf{Symbol}&\textbf{Meaning}\\
\hline
% \midrule
$q$ & Questions posed to the LLM \\

$g$& The reply generated by the LLM based on the question \\

$t_j$& The token generated in the $j$-th step of the LLM\\

$f_i$& The $i$-th fact can be decomposed\\

$\phi$& The relationship between $f_i$ and $t_j$\\

$U$& The uncertainty quantification function\\

$\theta$& Threshold for distinguishing high and low uncertainty values \\

$F$& The set of facts decomposed by $g$ \\
% \hline
\bottomrule
\end{tabular}
\label{tab1}
\end{center}
\end{table}

In the uncertainty quantification of multi-fact generation in LLMs based on trap questions, let the question be $q$, and let the output consist of $m$ tokens generated by the LLM be $g = \{t_1, t_2,..., t_m\}$, The generation of an LLM $g$ can be decomposed into $n$ facts $F$ through external models or methods, that is, $F = \{f_1, f_2,..., f_n\}$. Therefore, there is a mapping relationship $\phi$ between each fact $f_i$ and the token generated by the LLM, as expressed in (\ref{eq1}), where the token generated at the $j$-th step is denoted as $t_j$.
\begin{equation}
\phi: f_i \mapsto \{ t_j \in g \mid \text{token } t_j \text{ is relevant to } f_i \}.\label{eq1}
\end{equation}

Assuming the uncertainty score is denoted as $U$. In this scenario, it is necessary to find a suitable function $U$ that satisfies the relationship as shown in (\ref{eq2}), where the value of $\theta$ can be taken according to the actual situation.
\begin{equation}
\begin{cases}
U(g)>\theta, & \text{if } g \text{ is generated} \\
U(g)\leq\theta, & \text{if } g \text{ is refused to generate}
\end{cases}\label{eq2}
\end{equation}

In response to the issue of erroneous generations by the model, we seek to identify an appropriate function $U$ such that the more facts in $F$, the higher the uncertainty quantification function value of $U$, as expressed in (\ref{eq3}), where $|F|$ denotes the size of fact set $F$.

\begin{equation}
\begin{gathered}
U: F \rightarrow \mathbb{R}, \\
s.t.\forall F_1,F_2, \ |F_1| > |F_2| \implies U(F_1) > U(F_2).
\end{gathered}\label{eq3}
\end{equation}

\subsection{MulFactTrap Dataset Construction}\label{IIIB}
Existing multi-fact generation datasets have limitations in reflecting the robustness of LLMs when measuring uncertainty, and they also fail to comprehensively evaluate the performance of uncertainty estimation methods. Therefore, we have constructed a new multi-fact generation dataset that includes trap questions. This dataset can reflect the robustness of LLMs based on their generation performance on it and can also be used for uncertainty estimation to measure the robustness of uncertainty estimation methods. 

Large language models typically generate multiple facts when addressing biographical questions. For example, the statement ``Albert Einstein was a German-born theoretical physicist who is widely regarded as one of the most influential scientists of the 20th century." can be broken down into several distinct facts, such as ``Albert Einstein was a person born in Germany" and ``Albert Einstein was a theoretical physicist." Therefore, we primarily focus on the biographical generation scenario to construct our trap-question multi-fact generation dataset.

\begin{figure}[htbp]
\centerline{\includegraphics[width=0.5\textwidth]{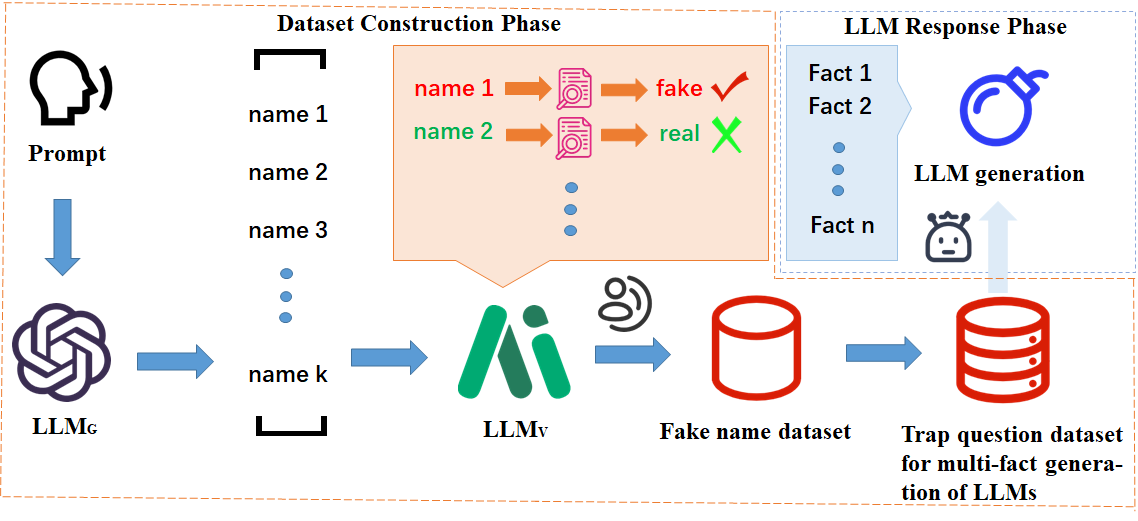}}
\caption{Overall framework for constructing our dataset.}
\label{fig2}
\end{figure}

First, we introduce a pipeline for constructing trap questions using LLMs, as illustrated in Fig.~\ref{fig2}. The pipeline consists of two main components: the LLM Generator ($LLM_G$), the LLM Verifier ($LLM_V$), LLMs used for generating and seperating facts. The $LLM_G$ is responsible for creating potential fake names, while the $LLM_V$ assesses the authenticity of the names generated by the LLM Generator.

Large language models have been empirically proven to possess the capability of generating biographies. This ability is primarily due to the extensive information about real individuals that they are exposed to during the training phase. However, when prompted to generate a biography of a person whose existence is uncertain during the inference stage, the likelihood of the model producing hallucinations significantly increases. To comprehensively evaluate the robustness of LLMs, we aim to construct pseudonyms that closely resemble real names. To achieve this, we employ the following three heuristic strategies for generating fake names: permutation and recombination, word fine-tuning, and fictional world creation.

\begin{itemize}
\item \textbf{Permutation and Recombination (PR):} The process involves splitting the segments of words in a name and recombining different segments from different names at the same positions. For example, after transformation, the names ``Donald John Trump" and ``Joseph Robinette Biden Jr." would be recombined to form the new name ``Donald Robinette Biden Jr.". 
\item \textbf{Word fine-tuning (WF):} Alter and fine-tune some letters in a person's name but ensure that the pronunciation does not change significantly after the modification. For example, ``Albert Einstan".
\item \textbf{Fictional world creation (FWC):} Incorporate character names from virtual worlds such as video games or anime. For example, ``Chiikawa".
\end{itemize}

The pipeline for constructing fake names consists of three steps: fake name generation, model verification of fake names, and manual fine-tuning of fake names. First, we manually construct prompts to instruct the $LLM_G$ to produce a list of $k$ potential fake names ($L_{pf}$) based on a series of real names and perturbation strategies. Since there may be overlaps between character names from fictional worlds and real names, we only use the first two heuristic strategies when generating names through the $LLM_G$.
Given the potential hallucination issues of the LLM Generator, the generated names in $L_{pf}$ may not be truly fake. Therefore, we conduct authenticity verification by querying the $LLM_V$ with the question, ``Is [name] a real person?" We select the names verified as “fake” to form a more refined list of potential fake names ($L_{mpf}$). Let the outcome of the t-th verification pass for the j-th instance be denoted by $c_{t}^{j} \in \mathcal{C}$. After T independent passes, the final verdict $c^{j}$ is taken as the majority vote over $\{c_1^j, …, c_T^j\}$. Ties trigger an additional round of verification until a clear majority emerges. Only instances whose ultimate verdict is “false” are retained for trap-question construction.
Finally, considering the possible hallucinations from the LLM Verifier, we perform manual verification and fine-tuning to ensure the quality of the fake names in the final dataset. We also manually exclude names that have significant pronunciation differences from the original real names and add some character names from fictional worlds using the fictional world creation strategy. After completing all these steps, we obtain the final dataset of fake names.

Specifically, we obtained our fake name dataset using data collected from FActScore\cite{b8} and the real biographical data used in the experiments of CCP\cite{b9}. We randomly selected 100 questions from these sources, extracted the names within them, and prompted Yi-Lightning to generate 100 potential fake names. After filtering, we had a pool of 80 fake names. Through manual review and minor adjustments, we refined this pool to 77 fake names for our experiments.

We conducted fact-checking on 100 potentially fake names generated by the Yi-Lightning model through single sampling. The models used for name-checking included Yi-Lightning\footnote{https://platform.lingyiwanwu.com/chat}, GPT-4o\footnote{https://chatgpt.com/}, and Kimi\footnote{https://kimi.moonshot.cn/}. The checking results are presented in Table~\ref{tab2}.

\begin{table}[htbp]
\caption{Name checking performance of 3 models}
\begin{center}
\renewcommand{\arraystretch}{1.5}
\begin{tabular}{>{\centering\arraybackslash}m{2cm}|>{\centering\arraybackslash}m{1.5cm}| >{\centering\arraybackslash}m{1cm}| >{\centering\arraybackslash}m{2cm}}
\toprule
% \hline
\textbf{Model}&\textbf{Accuracy}&\textbf{Recall}&\textbf{F1 score} \\
\hline
% \midrule
Yi-Lightning& 0.76 & 0.9091 & 0.8537 \\

GPT-4o& 0.7879 & 0.8947 & 0.8662 \\

Kimi& 0.7551 & 0.8831 & 0.85\\
% \hline
\bottomrule
\end{tabular}
\label{tab2}
\end{center}
\end{table}

As illustrated in Table~\ref{tab2}, all three models—Yi-Lightning, GPT-4o, and Kimi—performed well in the task of verifying fake names. The average accuracy of the labels provided by the models was 0.77, with an average recall rate of 0.90 and an average F1 score of 0.85. While GPT-4o achieved the best performance among the three, its use was constrained by the limitations on token numbers and the high inference costs associated with both GPT-4o and Kimi. Therefore, when constructing our dataset, we opted for the cost-effective and impressively performing Yi-Lightning model to conduct the verification of fake names.

Then, based on these names, we added the phrase ``Tell me a bio of" or ``Tell me a brief introduction of" in front of the fictional names to obtain the trap questions. The resulting trap question dataset can be used to evaluate the multi-fact generation capability of LLMs and measure their uncertainty. In this case, the probability of LLMs generating hallucinations on this dataset is significantly higher than on typical datasets. 

\begin{figure}[htbp]
\centerline{\includegraphics[width=0.5\textwidth]{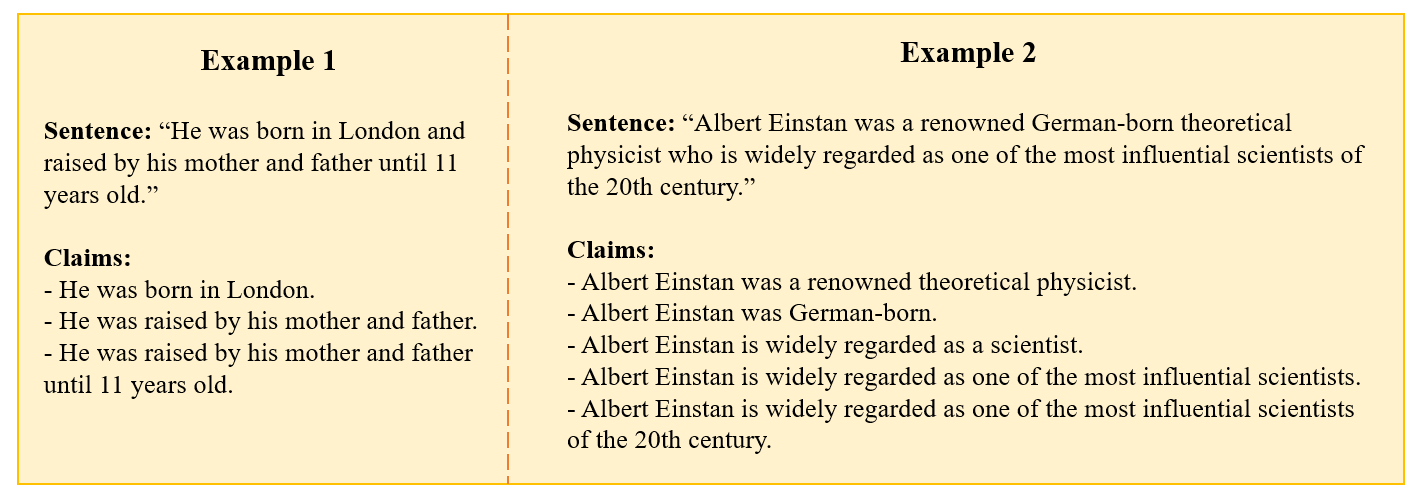}}
\caption{Two examples of the fact decomposition process.}
\label{fig3}
\end{figure}

Finally, during the LLM generation phase, we input each question from the dataset into the LLM or its API to be evaluated for generation uncertainty. By using the constructed trap questions to induce hallucinations in the LLM, and then decomposing the model's generated output into facts using a well-performing model such as GPT-4, we can obtain a trap question dataset for LLMs containing multiple facts and the generated texts. MulFactTrap comprises 77 meticulously crafted trap questions for the task of generating false biographies, with 35 percent constructed using the PR strategy, 58 percent using the WF strategy, and 7 percent using the WFC strategy. Additionally, MulFactTrap includes 385 model-generated data samples collected from four LLMs. During the process of fact decomposition, we employed a two-shot prompt. The first shot guided the LLM on how to perform fact decomposition, while the second shot instructed the model not to alter the false and erroneous facts in the original generation during the decomposition process. The two samples are shown in the Fig.~\ref{fig3}.

\subsection{Robust Uncertainty Quantification Method}

To better adapt to the uncertainty measurement scenario of LLM-generations for trap questions, we propose a robust uncertainty measurement method, termed $RU$. This method achieves more robust and accurate uncertainty quantification by decomposing and aggregating fact-level uncertainties in a fine-grained manner according to categories. The framework of the $RU$ method is illustrated in Fig.~\ref{fig4}.

\begin{figure*}[htbp]
\centerline{\includegraphics[width=1.0\textwidth]{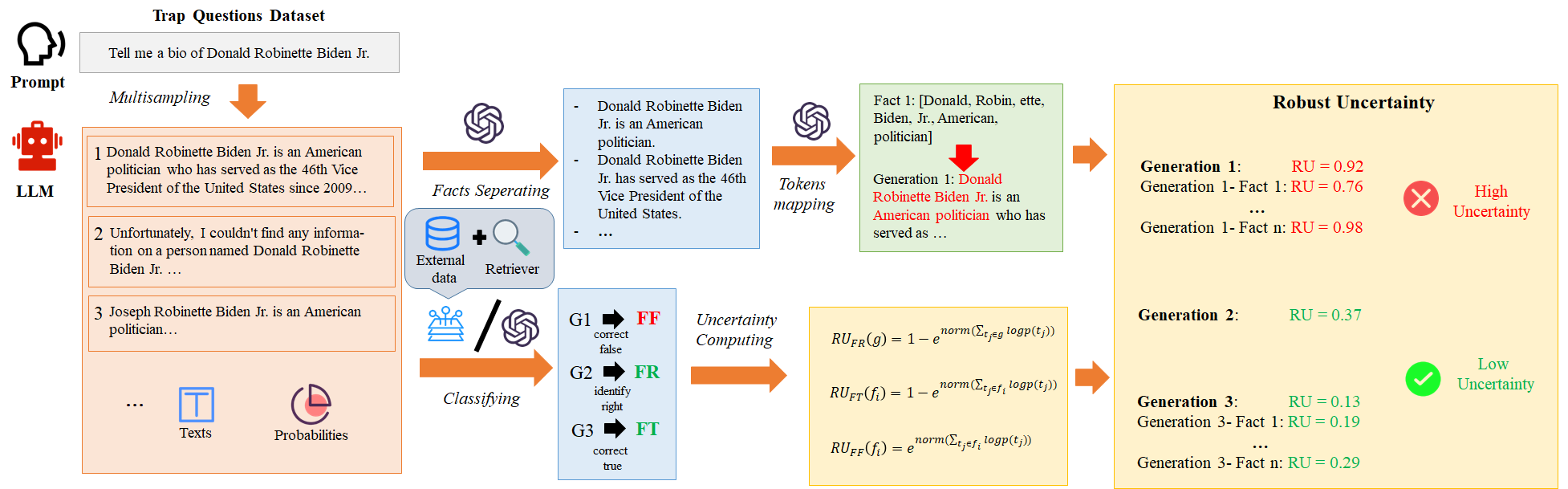}}
\caption{Overall framework for proposed $RU$ method.}
\label{fig4}
\end{figure*}

We initiate the process of obtaining the uncertainty score using RU by employing a multi-sampling strategy, the purpose of which is to achieve rich generation and semantic diversity. The primary reason is that multi-sampling not only comprehensively captures the potential semantic information of the model's generation but also allows us to sample a larger volume of generated data, even when the number of questions is limited. This approach enables more accurate evaluation of the uncertainty measurement performance. In our approach to multi-sampling, we adopt a combination of random sampling and top-k sampling strategies. The value of k is maintained at the model's default setting, and the sampling temperature is kept constant. We utilize beam search as the primary decoding strategy, with the maximum number of generated tokens set to 100.

After collecting the model's generations, we categorize the model's outputs into three types: Fact identify Right (FR), Fact correct True (FT), and Fact correct False (FF). FR indicates that the LLM correctly identifies errors or false content in the question and refuses to answer. FT denotes that the LLM fails to recognize the errors but corrects them to factual information during the generation process. FF signifies that the LLM fails to identify the false information and either leaves it uncorrected or replaces it with another fake or incorrect content. The classification labels can be obtained by constructing prompts to guide the LLMs, such as GPT-4 or Yi-Lightning, a method that is relatively universal. In the context of multi-fact generation, the selection of classification algorithms should be tailored to the specific characteristics of the generated content. Moreover, incorporating a retriever to search for corroborating facts within external datasets can enhance the verification process of the generated information. For example, in the uncertainty quantification of LLM's generation based on fake name trap questions, we first use the NLI model such as deberta to judge the logical relationship with the expression ``Unfortunately, I can't provide the information of ..." according to the generated text. If the result is entailment, it is marked as ``FR". Using Wikidata as external data, we use a retriever based on the BM25 retrieval algorithm to search Wikidata. If the description of the relevant person is retrieved, we mark the label as ``FT", otherwise, we classify it as ``FF".

In decomposing the model's generation into facts and mapping tokens, we follow the approach of prior studies by using prompts to guide the API of a high-performance LLM Yi-Lightning to complete this task. Fact decomposition, akin to the methods previously introduced, utilizes a two-shot prompting strategy as well. During the token mapping process, we input the entire generation and the individual facts obtained from decomposition into the Yi-Lightning prompt LLM to list the words in the original generation corresponding to each fact. Subsequently, we use the tokenizer of the original model to match the words with tokens and obtain their positions in the original generation. Finally, we obtain a list of the positions of all tokens corresponding to each fact in the original generation. Finally, we calculate the RU results based on the logits information from the LLM's generation, the category labels, and the mapping relationships between facts and generations.

We continue with the scenario proposed in Section \ref{IIIA}, assuming the model's generation is denoted as $g=\{t_1, t_2,..., t_m\}$. Due to the autoregressive generation of LLMs, the probability of the token generated at the $j$-th step is denoted as $p(t_j|t_{<j})$$(i=1,2,...,m)$. For the convenience of calculation and analysis, we approximate and simplify this probability as $p(t_j)$. We calculate the uncertainty of the generation using (\ref{eq4}) for outputs labeled as FR. For outputs labeled as FT, we compute the fact-level uncertainty using (\ref{eq5}), and compute for outputs labeled as FF using (\ref{eq6}), where $norm(.)$ denotes length normalization.

\begin{equation}
RU_{FR}(g) = 1-e^{norm(\sum_{t_j\in{g}}{logp(t_j)})},\label{eq4}
\end{equation}

\begin{equation}
RU_{FT}(f_i) = 1-e^{norm(\sum_{t_j\in{\phi(f_i)}}{logp(t_j)})},\label{eq5}
\end{equation}

\begin{equation}
RU_{FF}(f_i) = e^{norm(\sum_{t_j\in{\phi(f_i)}}{logp(t_j)})}.\label{eq6}
\end{equation}

\section{Experiment and results}\label{V}

In this section, we focus on the uncertainty quantification of biographical text generation involving fictional individuals. We will introduce the experimental setup for uncertainty quantification based on trap questions(Section \ref{VA}), and the results of the experiments along with their analysis(Section \ref{VB}).

\subsection{Experimental Settings}\label{VA}
\subsubsection{Data}
We conducted experiments on the uncertainty quantification of outputs generated by LLMs using MulFactTrap dataset introduced in Section \ref{IIIB}.

\subsubsection{Models}
We conducted experiments on uncertainty measurement using the following four LLMs: LLaMA3-8B-Instruct\cite{b27}, Vicuna-13B\cite{b28}, ChatGLM3-6B-32K\cite{b29}, and Mistral-7B\cite{b30}.

\subsubsection{Baseline methods}
We primarily employed the following four methods as the baseline approaches for RU: PE\cite{b14}, LN-PE\cite{b31}, Max prob\cite{b9} and CCP\cite{b9}. Since the Max prob and CCP methods are based on the fact granularity in \cite{b9}, we used the average and maximum values of uncertainty computed at the fact granularity to represent the uncertainty at the generation granularity.

\textbf{Predictive Entropy:} We employ the conventional predictive entropy as our baseline. Specifically, we denote the probability of the $j$-th generated token in the entire generation $g$ as $p(t_j|t_{<j})$. The predictive entropy is defined as the entropy value of the entire generation process. 

\begin{equation}
PE(g) = \sum_{t_j\in g}{p(t_j|t_{<j})logp(t_j|t_{<j})}.
\end{equation}

\textbf{Length-Normalized Predictive Entropy:} By normalizing the predictive entropy at the generation granularity based on length, we derive the Length-Normalized Predictive Entropy (LNPE). 

\begin{equation}
LN-PE(g) = norm(\sum_{t_j\in g}{p(t_j|t_{<j})logp(t_j|t_{<j})}).
\end{equation}

\textbf{Maximum Probability:} We simply regard the most probable generation probability as the confidence score and aggregate it at the fact granularity level. 

\begin{equation}
MP(f_i) = 1-\prod_{t_j\in f_i}p(t_j|t_{<j}).
\end{equation}

\textbf{Claim Conditioned Probability:} The CCP method first employs an NLI model\footnote{https://huggingface.co/microsoft/deberta-large-mnli} to compute the probabilities of words in the model's vocabulary that have entailment and contradiction relationships with the original word. This process yields word-level CCP scores. Subsequently, the product of these word-level CCP scores is calculated to obtain the claim-level CCP. 

\begin{equation}
CCP_{word}(t_j)=\frac{\sum_{k:\mathsf{NLI}(t_j^k,t_j)=\text{`}\mathsf{e}\text{'}}P(t_j^k\mid t_{<j})}{\sum_{k:\mathsf{NLI}(t_j^k,t_j)\in\{\text{`}\mathsf{e}\text{'},\text{`}\mathsf{c}\text{'}\}}P(t_j^k\mid t_{<j})},
\end{equation}

\begin{equation}
CCP_{claim}(f_i)=1-\prod_{t_j\in f_i}CCP_{word}(t_j).
\end{equation}

\subsubsection{Evaluation metrics}
We employ accuracy, recall and F1 score as evaluation metrics for the fact-checking step of our model, where the ground-truth labels are obtained through manual search. Consistent with prior research, we utilize the ROC-AUC score (ROC), pearson correlation coefficient (PC) and spearman correlation coefficient (SC) as the metrics for the uncertainty quantification. Currently, there is no comprehensive evaluation criterion to measure the robustness of the LLM. Therefore, we obtain the labels of correctness by prompting Yi-Lightning to determine whether there are hallucinations in the generation.

\subsection{Results and Analysis}\label{VB}

\subsubsection{Uncertainty quantification}
We employed four models to conduct multi-sampling based on the constructed dataset, with a sampling size of 5. During the generation process, the beam size was also set to 5 and the sampling temperature was set to 1.0. For each generated sample, we obtained the corresponding label through feature extraction. The Yi-Lightning model was utilized to perform fact decomposition and token mapping. The uncertainty evaluation results of the proposed $RU$ method compared to the baseline method are shown in Table~\ref{tab3}. Specifically, $RU_{gen}$ refers to aligning facts to the generation granularity, while $RU_{fact}$ indicates aligning generations to the fact granularity. $CCP_{max}$ and $Max prob_{max}$ represent the uncertainty of the entire generation based on the maximum uncertainty among all facts. $CCP_{mean}$ and $Max prob_{mean}$ denote the uncertainty of the entire generation based on the average uncertainty of all facts.

To more closely approximate the uncertainty measurement in real-world scenarios, we expanded the MulFactTrap dataset by incorporating 50 additional questions related to the generation of real biographies, resulting in a mixed dataset of 127 questions that include both real and fake biography generation tasks. Using this mixed dataset, we conducted uncertainty measurement experiments with four large language models. The performance of each method is shown in Table ~\ref{tab4}.

\begin{table*}[htbp]
\caption{The performance of RU and baseline methods on 4 models in MulFactTrap}
\begin{center}
\renewcommand{\arraystretch}{1.5}
\begin{tabular}{c|c|c|c|c|c|c|c|c|c|c|c|c}
% \hline
\toprule
Model&\multicolumn{3}{|c|}{LlaMA3-8B-Instruct}&\multicolumn{3}{|c|}{Vicuna-13B}&\multicolumn{3}{|c|}{ChatGLM3-6B-32K}&\multicolumn{3}{|c}{Mistral-7B}\\
\hline
Metric &ROC & PC & SC & ROC & PC & SC & ROC & PC & SC & ROC & PC & SC\\
\hline

$PE$\cite{b14}& 0.5353& 0.1053 & 0.0589 & 0.6798 & 0.2737 & 0.2911 & 0.6511 & 0.1407 & 0.1984 & 0.7469 & 0.4025 & 0.4194\\

$LN-PE$\cite{b31}& 0.5022 & 0.0561 & -0.0036 & 0.5268 & -0.0956
 & -0.0433 & 0.677 & 0.2042 & 0.2324 & 0.7467 & 0.4017 & 0.4191
\\

$CCP_{mean}$\cite{b9}& 0.7457 & 0.4196 & 0.4203 & 0.6835 & 0.3465 & 0.3178 & 0.8028 & 0.3622 & 0.393 & 0.7322 & 0.3879 & 0.3945\\

$CCP_{max}$\cite{b9}& 0.7301 & 0.3645 & 0.3936 & 0.6491 & 0.2418
 & 0.2582 & 0.7608 & 0.2872 & 0.3384 & 0.7811 & 0.4047 & 0.4775\\

$Max prob_{mean}$\cite{b9} & 0.735 & 0.4289 & 0.4021 & 0.7099 & 0.3459 & 0.3634 & \textbf{0.8558} & 0.3862 & \textbf{0.4617} & 0.7249 & 0.3451 & 0.3822\\

$Max prob_{max}$\cite{b9}& 0.7521 & 0.3957 & 0.4312 & 0.6606 & 0.2251 & 0.2781 & 0.8191 & 0.3658 & 0.4141 & 0.7106 & 0.2907 & 0.3577\\

\hline

$RU_{gen}$(ours)& 0.7442 & \textbf{0.5432} & 0.4075 & \textbf{0.8660} & \textbf{0.7314} & \textbf{0.5925} & 0.7775 & \textbf{0.4609} & 0.3644 & \textbf{0.8876} & \textbf{0.7868} & \textbf{0.6584}\\

$RU_{fact}$(ours)& \textbf{0.8195} & \textbf{0.6156} & \textbf{0.5267} & \textbf{0.8479} & \textbf{0.6998} & \textbf{0.5214} & 0.7647 & \textbf{0.4256} &0.3518
 & \textbf{0.898} & \textbf{0.7479} & \textbf{0.6684} \\

% \hline
\bottomrule
\end{tabular}
\label{tab3}
\end{center}
\end{table*}

\begin{table*}[htbp]
\caption{The performance of RU and baseline methods on 4 models in mixed dataset}
\begin{center}
\renewcommand{\arraystretch}{1.5}
\begin{tabular}{c|c|c|c|c|c|c|c|c|c|c|c|c}
% \hline
\toprule
Model&\multicolumn{3}{|c|}{LlaMA3-8B-Instruct}&\multicolumn{3}{|c|}{Vicuna-13B}&\multicolumn{3}{|c|}{ChatGLM3-6B-32K}&\multicolumn{3}{|c}{Mistral-7B}\\
\hline
Metric &ROC & PC & SC & ROC & PC & SC & ROC & PC & SC & ROC & PC & SC\\
\hline

$PE$\cite{b14}& 0.5093& 0.019 & -0.0141 & 0.6571 & 0.2674 & 0.2702 & 0.7359 & 0.3981 & 0.399 & 0.7052 & 0.3036 & 0.332\\

$LN-PE$\cite{b31}& 0.5141 & 0.0127 & -0.0214 & 0.6042 & 0.1476 & 0.1792 & 0.7366 & 0.3997 & 0.4002 & 0.7263 & 0.3608 & 0.3662\\

$CCP_{mean}$\cite{b9}& \textbf{0.8025} & 0.5283 & \textbf{0.5076} & 0.7119 & 0.3988 & 0.364 & 0.6774 & 0.3108 & 0.2835 & 0.6793 & 0.3061
 & 0.3106\\

$CCP_{max}$\cite{b9}& 0.7907 & 0.4591 & 0.4879 & 0.7183 & 0.3617 & 0.375 & 0.7354 & 0.3457 & 0.3762 & 0.7349 & 0.3465
 & 0.4068\\

$Max prob_{mean}$\cite{b9} & 0.7853 & \textbf{0.5486} & 0.4787 & 0.7467 & 0.4393 & 0.4239 & 0.7229 & 0.3519 & 0.3561 & 0.724
 & 0.3467 & 0.388\\

$Max prob_{max}$\cite{b9}& 0.7804 & 0.5237 & 0.4705 & 0.7226 & 0.3508 & 0.3825 & 0.7121 & 0.3512 & 0.3389 & 0.7413 & 0.3712 & 0.4179\\

\hline

$RU_{gen}$(ours)& 0.6905 & 0.4929 & 0.2908 & \textbf{0.9001} & \textbf{0.7941} & \textbf{0.6883} & \textbf{0.7636} & \textbf{0.4854} & \textbf{0.4232} & \textbf{0.8562} & \textbf{0.7304} & \textbf{0.6024}\\

$RU_{fact}$(ours)& 0.7393 & 0.5304 & 0.3495 & \textbf{0.899} & \textbf{0.7874} & \textbf{0.6882} & \textbf{0.7515} & \textbf{0.4347} &0.396 & \textbf{0.8753} & \textbf{0.7145} & \textbf{0.6394} \\

% \hline
\bottomrule
\end{tabular}
\label{tab4}
\end{center}
\end{table*}

Based on the results presented in Table~\ref{tab3} and Table~\ref{tab4}, we can draw the following conclusions: First, the proposed RU method outperforms all the compared baseline methods in the vast majority of cases. The average ROC-AUC value of RU is 0.1 to 0.2 higher than that of the baseline methods. Additionally, RU demonstrates significant advantages over the baseline methods in terms of both pearson correlation coefficient and spearman correlation coefficient. Second, among the four baseline methods, CCP and Max Prob generally exhibit better performance than the others. This indicates that fine-grained decomposition of facts and uncertainty measurement at the fact level can enhance the overall performance and robustness of uncertainty estimation. Finally, compared to CCP and Max Prob, our RU method further improves performance by incorporating additional classification and fine-grained calculations for different types of generations. This more detailed classification and calculation approach further enhances the performance of uncertainty estimation. The results show that both generation-level RU and fact-level RU maintain well-calibrated behavior under most perturbation scenarios. Baseline methods exhibit markedly lower correlation, with some even yielding negative values on the perturbed datasets, revealing their limited robustness to disturbances and adversarial attacks. In contrast, RU leverages fine-grained fact decomposition and uncertainty-aware classification to successfully detect trap questions and latent factual hallucinations in model generation.

\subsubsection{The robustness of LLMs for trap questions}
We have calculated the probability of generating ``hallucinations" (labeled as ``FF") in the MulFactTrap. The results are shown in Fig. 5.

\begin{figure}[htbp]
\centerline{\includegraphics[width=0.5\textwidth]{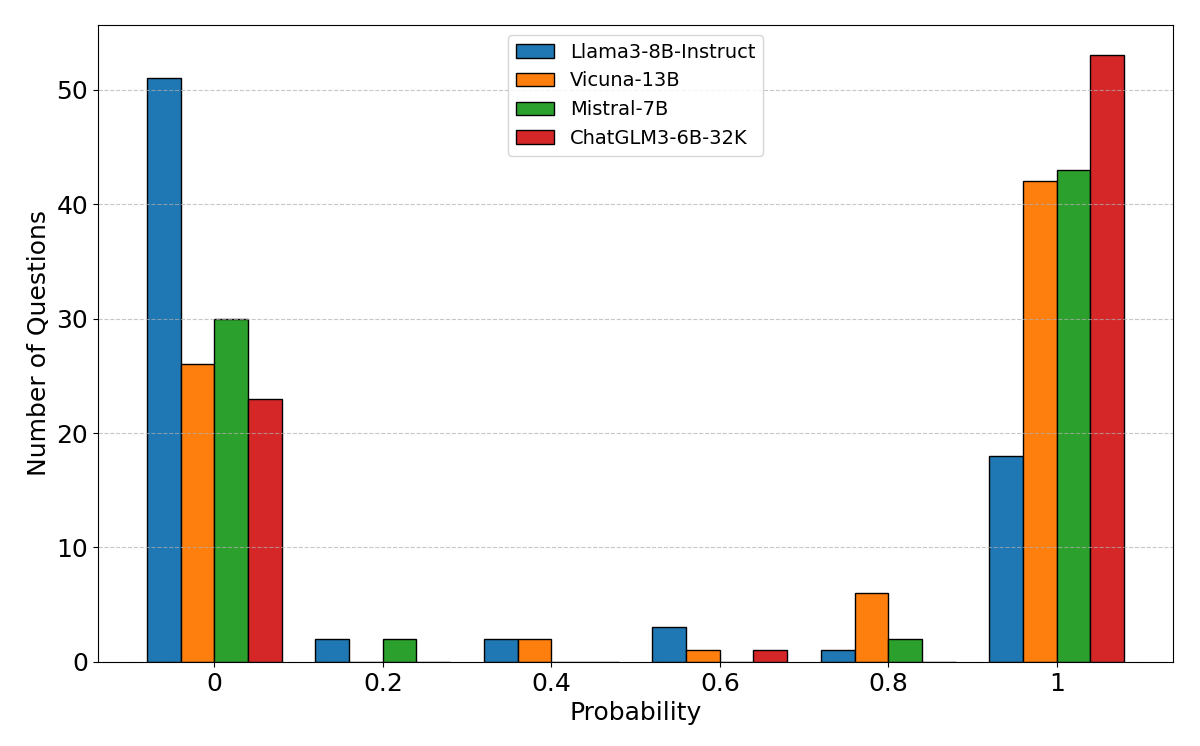}}
\caption{Probability distribution of hallucinations in model generations.}
\label{fig4}
\end{figure}

The distribution shown in Fig.~\ref{fig4} reveals a bimodal tendency in the probability of hallucinations when these models face fictional biographies. This suggests that when confronted with the same deceptive task, a single model is more likely to produce either entirely correct or entirely incorrect outputs. Among the different models, the Llama3-8b-instruct model exhibits a lower average probability of hallucinations, indicating better robustness. In contrast, the ChatGLM3-6B-32K model has a higher probability of hallucinations, suggesting a relatively weaker robustness. Our analysis leads to three main conclusions: The evaluated models are highly susceptible to hallucination when confronted with perturbed questions; Instruct-tuned variants exhibit lower hallucination rates than their base counterparts, an effect we attribute to alignment procedures that internalize latent human preferences present in the training data; Smaller models succumb to hallucination more readily than larger ones under perturbation.

\subsection{Ablation Study}
We conducted ablation studies on the task of verifying fictional names, focusing on the sampling size $k$ and whether to employ a Chain-of-Thought (CoT) strategy during the name verification process.

\subsubsection{The sampling size $k$}
We investigated the impact of different sampling sizes $k$ in the Yi-Lightning model on the performance of name verification, with the results presented in Table~\ref{tab4}. For cases involving multiple samples, we employed a majority voting strategy, selecting the label with the highest frequency in the generated domain as the final label.

\begin{table}[htbp]
\caption{Name checking results of different sample sizes}
\begin{center}
\renewcommand{\arraystretch}{1.5}
\begin{tabular}{c|c|c|c}
\toprule
% \hline
\textbf{Sample size}&\textbf{Accuracy}&\textbf{Recall}&\textbf{F1 score} \\
\hline
1 & 0.76 & 0.9091 & 0.8537 \\

3 & 0.77 & 0.9221 & 0.8606 \\

5 & 0.78 & 0.9351 & 0.8675\\

7 & 0.77 & 0.9221 & 0.8606\\
% \hline
\bottomrule
\end{tabular}
\label{tab5}
\end{center}
\end{table}

As shown in Table~\ref{tab5}, when the number of samples is relatively small, the performance of the person verification task gradually improves with the increase of the sample size $k$, albeit at a slow pace. The performance peaks at $k$ = 5, and further increases in $k$ do not lead to significant improvements. Given that larger sampling sizes require more time, we did not conduct experiments for larger values of $k$.

\subsubsection{CoT strategy}

We conducted experiments to assess the verification performance of the Yi-Lightning model with and without the Chain of Thought (CoT) strategy when the sample size was set to 3. When the CoT strategy was not applied, the model's verification accuracy decreased from 0.77 to 0.65, recall dropped from 0.9221 to 0.7143, and the F1 score correspondingly declined from 0.8606 to 0.7586. These findings clearly indicate that the CoT strategy substantially enhances the model's verification capability.

\section{Conclusion and future work}
We introduce a novel task scenario for uncertainty quantification with trap questions based on multi-fact generation and provide a formal definition for it. This scenario extends previous work by simultaneously considering the robustness of the model. We propose a pipeline for constructing trap questions dataset MulFactTrap to implement the uncertainty quantification scenario and validate its effectiveness through experiments. Additionally, we present a new robust method RU for uncertainty measurement, demonstrating its superior performance on four models. Our method achieves an approximate 0.1-0.2 improvement in metrics compared to the current state-of-the-art approaches.

While the methods and experimental results of this work are satisfactory, several limitations may still exist. First, the uncertainty measurement method proposed in this paper, RU, has not yet been validated for application in other trap problem scenarios. The transferability of this method may be subject to limitations. Second, the robustness of the models studied in this work when facing trap problems may not be comprehensive enough. A more holistic evaluation criterion is needed to assess the robustness of the models. Finally, By leveraging the pipeline constructed with the proposed dataset and incorporating a greater number of real names, the trap dataset in this paper can achieve good scalability. In the future, the scale of the dataset can be further expanded to better serve downstream NLP tasks. Considering these aspects may further enhance the performance of uncertainty quantification.

% \section*{References}
\section*{Acknowledgment}

This work was supported in part by the National Key Research and Development Program of China under Grant 2023YFC3305402, and in part by the National Natural Science Foundation of China (Nos.62302059 and 62172053).
We are also grateful to Qinhong Lin, a fellow PhD student in our research group, for his valuable suggestions on the initial draft of this manuscript and for engaging in insightful discussions regarding the experiments.

\vspace{12pt}

% \begin{appendices}
% \section{prompt}

% \section{bbb}
% \end{appendices}

\end{document}